%% file: IEEE_Picerno_RiL.tex
\documentclass[lettersize,journal]{IEEEtran}
\usepackage{amsmath,amsfonts}
\usepackage{amssymb}
\usepackage{algorithmic}
\usepackage{algorithm}
\usepackage{array}
\usepackage{balance}
\usepackage{blindtext}
\usepackage{capt-of}
\usepackage[caption=false,font=normalsize,labelfont=sf,textfont=sf]{subfig}
\usepackage{textcomp}
\usepackage{stfloats}
\usepackage{commath}
\usepackage{footnote}
\usepackage{mathtools}
\usepackage{textcomp}
\usepackage{stfloats}
\usepackage{multirow}
\usepackage{multicol}
\usepackage{float}
\usepackage{nomencl}
\usepackage[caption=false,font=normalsize,labelfont=sf,textfont=sf]{subfig}
\usepackage{svg}
\usepackage{url}
\usepackage{breakurl}

\usepackage{verbatim}
\usepackage{graphicx}
\usepackage{xcolor}
\usepackage[nocompress]{cite}
\usepackage{babel}
\usepackage[range-phrase={-},per-mode=symbol-or-fraction,inter-unit-product=,binary-units=true,range-units=single,list-units=single,detect-all,forbid-literal-units=true]{siunitx}
\DeclareSIUnit{\kph}{kph}
\usepackage{tabularx}
\usepackage{booktabs}
\usepackage{makecell}
\usepackage{caption}

\usepackage[switch]{lineno}

\input{acronyms}

\begin{document}
\title{Transfer of Reinforcement Learning-Based Controllers from Model- to Hardware-in-the-Loop}

\author{
   Mario Picerno,
   Lucas Koch,
   Kevin Badalian,
   Marius Wegener,
   Joschka Schaub,
   Charles Robert Koch,
   and Jakob Andert
    \IEEEcompsocitemizethanks{
        \IEEEcompsocthanksitem This work has been submitted to the IEEE for possible publication. Copyright may be transferred without notice, after which this version may no longer be accessible.
        \IEEEcompsocthanksitem This work was carried out in parts at the “Center for Mobile Propulsion (CMP)” of the RWTH Aachen University, funded by the German Research Foundation (DFG). This Project is supported by the Federal Ministry for Economic Affairs and Climate Action (BMWK) on the basis of a decision by the German Bundestag (KK5371001).
        \IEEEcompsocthanksitem M. Picerno, L. Koch, K. Badalian and J. Andert are with the Teaching and Research Area Mechatronics in Mobile Propulsion, Forckenbeckstraße 4, 52062 Aachen, Germany.
        \IEEEcompsocthanksitem M. Wegener and J. Schaub are with FEV Europe GmbH, Neuenhofstraße 181, 52080 Aachen, Germany.
        \IEEEcompsocthanksitem C. R. Koch is with the Department of Mechanical Engineering, University of Alberta, Edmonton, AB T6G 1H9, Canada.
        \IEEEcompsocthanksitem Corresponding author: Jakob Andert (e-mail: andert@mmp.rwth-aachen.de).
}}

\maketitle
\begin{abstract}
The process of developing control functions for embedded systems is resource-, time-, and data-intensive, often resulting in sub-optimal cost and solutions approaches.
Reinforcement Learning (RL) has great potential for autonomously training agents to perform complex control tasks with minimal human intervention.
Due to costly data generation and safety constraints, however, its application is mostly limited to purely simulated domains.
To use RL effectively in embedded system function development, the generated agents must be able to handle real-world applications.
In this context, this work focuses on accelerating the training process of RL agents by combining Transfer Learning (TL) and X-in-the-Loop (XiL) simulation.
For the use case of transient exhaust gas re-circulation control for an internal combustion engine, use of a computationally cheap Model-in-the-Loop (MiL) simulation is made to select a suitable algorithm, fine-tune hyperparameters, and finally train candidate agents for the transfer.
These pre-trained RL agents are then fine-tuned in a Hardware-in-the-Loop (HiL) system via TL.
The transfer revealed the need for adjusting the reward parameters when advancing to real hardware. 
Further, the comparison between a purely HiL-trained and a transferred agent showed a reduction of training time by a factor of \num{5.9}.
The results emphasize the necessity to train RL agents with real hardware, and demonstrate that the maturity of the transferred policies affects both training time and performance, highlighting the strong synergies between TL and XiL simulation.
\end{abstract}


\section{Introduction}
\label{sec:introduction}
\IEEEPARstart{A}s the automotive industry embarks on a transition toward software-defined and electrified transportation, the optimization of powertrain control represents a crucial step towards a sustainable future.
Embedded systems must effectively operate multiple domains to meet stringent performance requirements, but their development requires engineers and specialists in essential tasks, including design, calibration, and validation.
The automation of product development, achieved by assisting human-driven tasks with \ac{ML} methods, presents an intriguing avenue worth exploring to potentially expedite the technology transfer process \cite{DWIVEDI2021101994}.

\ac{RL} has gained considerable attention due to its potential to learn complex control strategies automatically.
In fact, the inclusion of \ac{RL} algorithms not only reduces the burden of domain experts, but also satisfies the rising demand for hands-free solutions.
To make effective use of \ac{RL} in practical development tasks, however, methods and tools are required to generate agents that can meet the demands of real-world operation.
Further, data generation is getting increasingly costly when training with physical prototypes, so the need for an efficient procedure to leverage existing knowledge becomes essential to reduce training demands.

\ac{RL} has shown relevance in diverse powertrain-related tasks, like emission and energy management, limited to training and validation solely within simulations \cite{Bae2020UreaIC,Tang21,deFrahan2021,Norouzi2023}.
The application of \ac{RL} in real-world environments is still limited due to data inefficiency, limited exploration capabilities, and hardware constraints \cite{DulacArnold2019}. Safe and constrained \ac{RL} techniques are being investigated also to address risk-mitigated learning processes in complex and dynamic environments \cite{osti_10313521,gu2023review}.

Moreover, \ac{TL} has emerged as a powerful technique, allowing models to transfer knowledge to a target domain to enhance learning and performance without the necessity of extensive training in the target domain \cite{SurveyTL, SurveyTLRL}. 
In \ac{RL}, this technique is adopted with a simulation model representing the source domain and the real-world application defining the target domain, referred to as \emph{sim-to-real} \cite{Zhao20Sim2Real}
While sim-to-real experiments are common in robotics \cite{liu2022digital,ju2022transferring}, the application of \ac{RL} to real-world powertrain control must face the challenges of modelling non-stationary and highly complex processes \cite{1470476}. This poses significant obstacles to a seamless transfer to real-world systems.
Model-free \ac{RL} agents for predictive energy management of a hybrid vehicle are trained in a \ac{HiL} test bed in \cite{Zhou2019} to evaluate real-time capability.
In \cite{BOOKT9376968}, an \ac{RL}-based controller is pre-trained offline and then successfully deployed in a real electric motor, indicating a decent performance on the target domain.
Nevertheless, the need to improve a policy using \ac{TL} techniques inside progressively complex systems is a documented problem \cite{Balakrishnan2021,Che9343713}.

\emph{In-the-Loop} approaches encompass a range of model-based simulation techniques that integrate different versions of a physical system into a closed-loop simulation \cite{Grossmann2011model,Fagcang2022}.
In automotive applications, the benefits of model-based \ac{XiL} methods lie in their ability to develop embedded systems within realistic environments \cite{TIBBA2016}. Here, \ac{HiL} simulations are frequently employed for the virtual calibration of \acp{ECU} \cite{Lee2018a,Gottorf2021,Dorscheidt2021,Riccio22}. 
\ac{RL} and \ac{XiL} virtualization share a need for precise and computationally efficient models of the interacting physical components.
By integrating real hardware, the inevitable disparity between simulation and real-world conditions can be reduced, supporting the adaptation of \ac{RL} agents to real-world scenarios. 
Thereby, the step-wise progression through virtualization stages throughout the \ac{XiL}-based development process could go hand in hand with transferring the agent's policy between different environments.
Thus, \ac{XiL} can serve as a robust and cost-effective foundation for \ac{RL}-based control function development, allowing for a rapid transition into real-world applications.

There is a shortage of research on data-efficient transfer procedures for powertrain control development pertinent to real-world scenarios.
Our contribution aims at closing this research gap by analyzing the transfer capabilities of \ac{RL} agents between \ac{XiL} stages.
To the best of our knowledge, this study represents the first \ac{TL} approach that performs training in the real-world domain in the context of powertrain control.

Our previous study demonstrated the ability of \ac{RL} agents to learn effective policies for emission-related control tasks in a \ac{MiL} environment.
For transient engine \ac{EGR} control, training utilized driving cycles, with policy validation across various maneuvers and ambient scenarios \cite{KOCH2023105477}.
On this basis, two state-of-the-art \ac{RL} algorithms are tested in the \ac{MiL} stage, and selected agents with different maturity levels are transferred and further trained with real hardware.
A \ac{HiL} platform equipped with a real \ac{ECU} and real actuators can replicate the challenges presented by real hardware, while also providing a secure environment for the agent to learn without risking hardware damage. 
For seamless execution and fine-tuning of pre-trained agents on the \ac{HiL}, we make use of the \ac{XiL}-based training of \ac{RL} agents established in our previous study and showcase its applicability in a use case where both virtual and real environments are present.
Finally, the goal of this study is to explore a feasible transfer procedure that yields the best trade-off between performance and training time.

This article is structured as follows: Sec. \ref{sec:Methodology} briefly introduces the \ac{RL} methods and algorithms used for this study as well as the \ac{XiL} toolchain that constitutes the basis of the study.
In Sec.~\ref{sec:RL_application} the use case of transient \ac{EGR} control, the application of \ac{RL} to the former and the utilized toolchain are presented.
A comparison between two \ac{RL} algorithms results in the simulated domain and an analysis of different transfers on the \ac{HiL} are discussed in Sec.~\ref{sec:Results}. Finally, the conclusions are summarized.

\section{Methodology}
\label{sec:Methodology}
\subsection{Reinforcement Learning}
\label{subsec:reinforcement-learning}
\Ac{RL} is a \ac{ML} paradigm that rests on free interactions between an agent and an environment to optimize the agent's strategy.
The framework for the environment is a discrete-time \emph{Markov Decision Process}.
It is formulated as a tuple $(S, A, P, R)$ which is made up of the state space $S$, 
the action space $A$, a function $P: S \times A \times S \to [0, 1]$ that determines the probabilities of state transitions, and a reward function $R: S \times A \times S \to \mathbb{R}$.
The agent, on the other hand, is defined by its policy $\pi_{\theta}: S \times A \to [0, 1]$, an action distribution with mutable parameters $\theta$.
The policy is generally represented by a \ac{NN}, meaning that $\theta$ contains its 
weights and biases. \cite{Sutton98}

In every time step, a single interaction takes place between the two entities: The agent observes the current state $s_{t} \in S$ and picks an action $(a_{t} \sim \pi_{\theta}({} \cdot {} | s_{t})) \in A$.
As a consequence, the environment proceeds to the next state $(s_{t + 1} \sim P({} 
\cdot {} | s_{t}, a_{t})) \in S$ and gives back a reward $r_{t} = R(s_{t}, a_{t}, s_{t + 1})$.
Interactions are recorded as tuples $(s_{t}, a_{t}, s_{t + 1}, r_{t})$ known as experiences.
A sequence $\tau$ of experiences starting in an initial and ending in a terminal state is called an episode.
The (discounted) sum of all rewards $R(\tau) = \sum_{t = 0}^{T}{\gamma^{t} r_{t}}$ for $\gamma \in (0, 1]$ is its return. \cite{Sutton98}.

\ac{RL} works towards maximizing the expected return $J(\pi_{\theta}) = E_{\tau \sim \pi_{\theta}}[R(\tau)]$ by incrementally updating $\theta$ via gradient ascent or some other suitable method.
To do so, \ac{RL} algorithms often internally make use of the value function $V_{\pi_{\theta}}(s) = E_{\tau \sim \pi_{\theta}}[R(\tau) | s_{0} = s]$, the action-value function $Q_{\pi_{\theta}}(s, a) = E_{\tau \sim \pi_{\theta}}[R(\tau) | s_{0} = s, a_{0} = a]$, or the advantage function $A_{\pi_{\theta}}(s, a) = Q_{\pi_{\theta}}(s, a) - V_{\pi_{\theta}}(s)$. \cite{Sutton98}

\subsubsection{Transfer Learning}
Inspired by psychological theories, \ac{TL} aims to apply knowledge gained in one \ac{ML} problem to a different yet related one.
When successful, training with \ac{TL} also requires less data and, consequently, less time compared to training from scratch, leading models to achieve superior performance \cite{torrey2010transfer}.

A common differentiation involves the feature spaces within the tasks in the source and target domains, distinguishing between homogeneous \ac{TL} with equal and heterogeneous \ac{TL} with differing feature spaces \cite{weiss2016survey}.
When investigating the knowledge aspect, one approach is to classify strategies according to how the policy is transferred to a new task. Direct parameter transfer, for instance, entails copying network parameters from a previous task to initiate training for a new task, whereas meta solution transfer combines previously learned policies into a meta-policy, condensing all prior knowledge for new tasks \cite{Glatt2023}.  

In the \ac{RL} context, \ac{TL} has the potential to address the challenges associated with training in fast and real-world environments.
Simulations constitute an environment where a vast amount of data can be collected without physical risk and then used to train \ac{RL} agents that are then deployed in the real use case (\emph{sim-to-real} \cite{Zhao20Sim2Real}).
A classification based on training samples required in the target domain distinguished between zero- and few-shot transfer, where no or only a few samples are needed in the target domain, and sample-efficient transfer, which requires a significant number of training samples from the target domain \cite{zhu2023transfer}.

\subsubsection{RL Algorithms}
\label{subsec:rl-algorithms}
\ac{RL} algorithms can be divided into different categories based on the qualities they possess.
In the context of this paper, the most pertinent characteristics revolve around access to an explicit environment model and the nature of the training data.
Certain \ac{RL} methods expect a model of the environment or learn one by themselves to predict the outcome of actions.
These \emph{model-based} approaches generally exhibit a high sample efficiency since the agent can use the model to plan its course of action and thereby require less exploration.
In contrast, \emph{model-free} algorithms do not rely on a pre-existing model of the environment but learn directly from interaction.
This has two core benefits in the context of \ac{RL}-based function development: For one, engineers are not burdened with the task of creating a sufficiently accurate environment model.
Furthermore, whether the environment models are human-made or learned, they can negatively affect the agent by introducing bias. \cite{Sutton98, spinningUpKindsOfRlAlgorithms}.

Another distinction is the type of data that \ac{RL} uses to learn.
Some require that the actions in the training data are sampled from the latest version of the agent's policy.
These \emph{on-policy} methods discard previous experiences and are therefore usually less sample efficient.
\emph{Off-policy} algorithms, on the other hand, use not only their own previous experiences, but they can also train on data from other entities, e.g. real-world data. \cite{Sutton98, spinningUpKindsOfRlAlgorithms}

In this work, two state-of-the-art \ac{RL} algorithms are deployed for the given control task.
The first is \ac{PPO} \cite{schulman2017proximal}, a model-free on-policy method capable of dealing with discrete and continuous action spaces.
Instead of looking at the expected return, it seeks to minimize a surrogate loss function so that advantageous actions are promoted while ensuring that the policy does not change drastically in a single update step. Following the actor-critic approach, the algorithm trains one \ac{NN} for the agent's policy and one to approximate the value function (cf. Sec. \ref{subsec:reinforcement-learning}).
Besides being more stable than comparable methods, \ac{PPO} is also fairly easy to configure because of its relatively low number of hyperparameters. \cite{schulman2017proximal, spinningUpIntroIntoPolicyOptimization}
The other algorithm, \ac{DDPG} \cite{lillicrap2019continuous}, likewise forgoes an explicit environment model, but differs in that it is off-policy and only supports
continuous action spaces.
\ac{DDPG} trains a \ac{NN} to act as an approximator of the Q-function (cf. Sec. \ref{subsec:reinforcement-learning}) and another \ac{NN} that yields the argument of the maximum of the same, making it an actor-critic technique as well.
To stabilize the process, the method makes use of time-delayed target networks when performing its calculations.
As a consequence of the above, \ac{DDPG} is harder to tune and more prone to instabilities than \ac{PPO} as it fulfills all criteria of the deadly triad of \ac{RL}: the algorithm utilizes function approximators (\acp{NN}, that is), it relies on bootstrapping (e.g. target networks that are themselves estimates), and it supports off-policy training. \cite{lillicrap2019continuous, spinningUpDDPG, Sutton98}

\subsection{X-in-the-Loop Toolchain}
At the core of \ac{XiL} simulations lies the flexibility to use simulated components across different levels of virtualization or use physical computing hardware (\ac{HiL}) in closed-loop.

\subsubsection{Real-time Models}
\label{sec:models}
The investigation is conducted using accurate models calibrated from a D-segment vehicle, and validated using engine test bench and chassis dynamometer test data.
The vehicle is equipped with a \SI{2.0}{\liter} compression ignition engine, with a single-stage turbocharger and \ac{HP} and \ac{LP} \ac{EGR} systems.
The powertrain consists of MATLAB/Simulink models for the vehicle, gearbox, and driver, as well as the physics-based real-time \ac{MVEM} and aftertreatment system. The models have been described in previous work \cite{KOCH2023105477}.
The physics-based method combines calibrated maps with physical equations for each component, including for the prediction of in-cylinder NO\textsubscript{x} and soot emissions.
Each function addresses variations in manifold pressures and temperatures, in addition to engine coolant temperature, and account for variations in ambient conditions and fuel type.
While the models are suitable for the virtual calibration of real embedded systems, they do not include the capability to simulate component aging or account for noise, vibration, and harshness factors. 

\subsubsection{Model-in-the-Loop and Hardware-in-the-Loop Frameworks}
The \ac{MiL} framework utilizes an air path controller emulating a real \ac{ECU} 
based on a model-based feed-forward control for \ac{EGR} and boost pressure with individual feedback control for model adaptation.
The controller determines the in-cylinder O\textsubscript{2} quantity to fulfill the levels of emissions \cite{Schaub2015RobustEC}.
The engine \ac{EGR} rate is regulated by converting an engine-out NO\textsubscript{x} set point into an in-cylinder O\textsubscript{2} concentration target.
Map-based control functions estimate the targets, while the connected plant models estimate the physical quantities.
To ensure the correct \ac{EGR} rate, a cascaded boost and delta pressures (exhaust and intake manifolds) control for both the turbocharger and the intake throttle ensures the necessary delta pressure.
The simulation can be executed in an accelerated time frame. 

\begin{figure}[ht]
    \centering
    \includegraphics[scale=0.35]{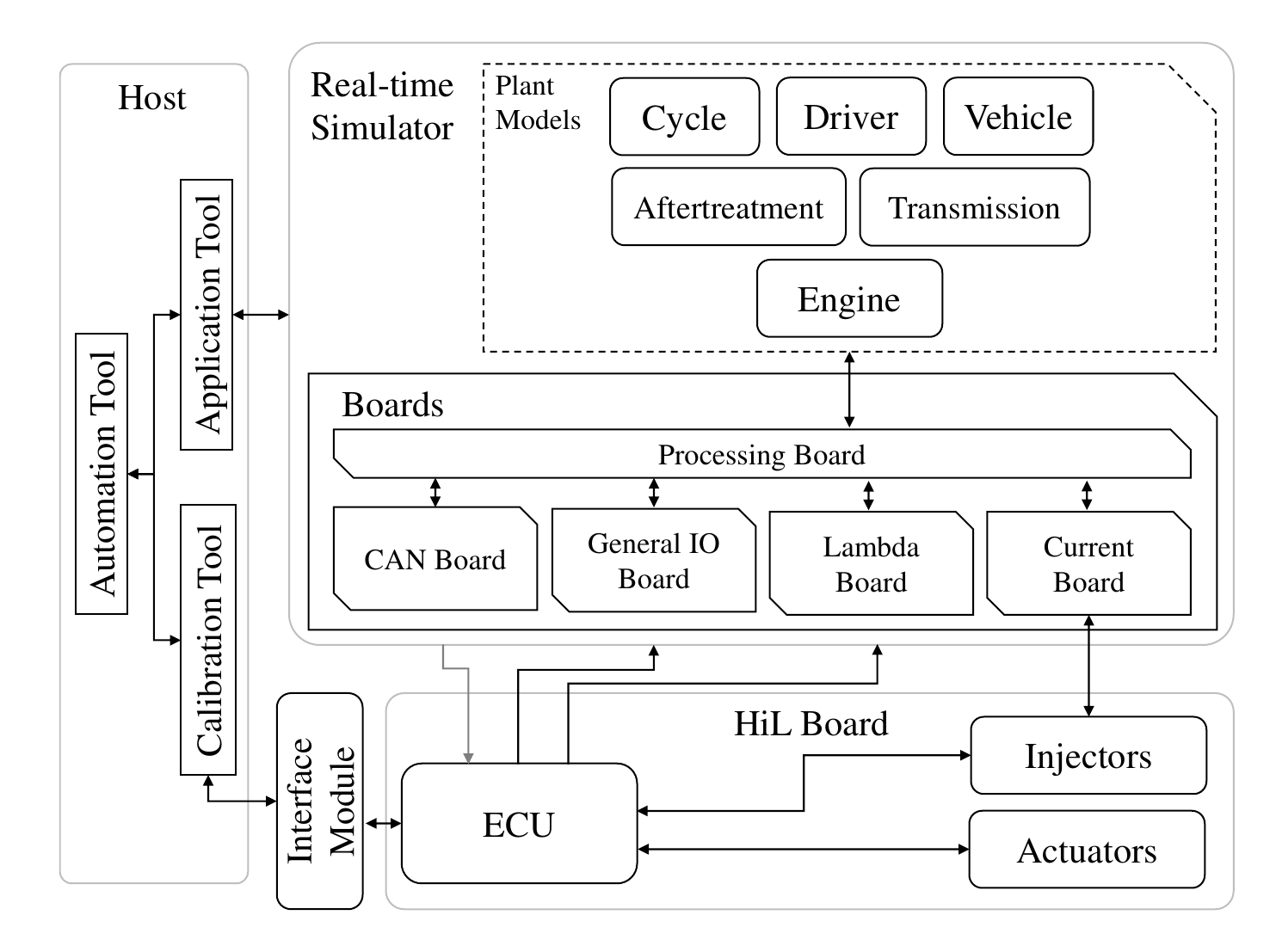}
    \caption{Schematics of \ac{HiL} system.}
    \label{fig:HiLtestbed}
\end{figure}

At the next level, the targeted \ac{HiL} system incorporates real hardware components, including a state-of-the-art production \ac{ECU}, turbocharger, throttle valves, and injectors, all placed on the \ac{HiL} board. These physical elements enable the functioning of the device under test.
The \ac{HiL} board is connected to a real-time simulator, where calculations are performed for system models, buses, protocols, and electrical interfaces (Fig. \ref{fig:HiLtestbed}).
The I/O boards of the simulator handle the electrical interfaces using digital/analog signals, resistive converters, and \ac{PWM} output channels to simulate sensor signals.
A dedicated board emulates the \ac{LSU}.
The \ac{HiL} board connects fuel injectors, from which current is measured and transformed into voltage to determine injection events.
The throttle plate is connected to the \ac{ECU}, and its position is measured in parallel by an Analog-to-Digital converter. The \ac{CAN} board enables the sending of essential \ac{CAN} signals, such as gear shifting and vehicle speed, to the \ac{ECU}.

Calibration software enables access to the embedded system on the \ac{HiL} board and an application software is responsible for the download and control of the plant models within the simulator.
Automation tools ensure real-time measuring and sensing between controllers and models, as well as automated testing.
The test bench provides a platform accurate enough for virtual \ac{RDE} calibration \cite{Lee2018b}.

\section{RL Application to Control Development}
\label{sec:RL_application}
This chapter describes the methodology's application, outlining the training workflow and the software architecture.

\subsection{Use Case of Air-path Management}
\label{subsec:problem_formulation}
Designing the \emph{Markov Decision Process} involves establishing states, actions, and rewards (Table \ref{tab:agentIO}).
The state space, specified by domain experts, must encompass all relevant information crucial for effective control of the \ac{EGR} valve.
These mainly include engine speed, pressures, and accelerator pedal position, along with their respective temporal deviations, which are utilized to identify process dynamics. Additionally, vehicle conditions such as coolant temperature, vehicle speed, and gear selection are taken into account.
For this study, the \ac{LP}-\ac{EGR} path was deactivated so that both the reference controllers in \ac{MiL} and \ac{HiL} as well as the agent control only the \ac{HP}-\ac{EGR} path.
All selected raw signals are min-max normalized to the interval $[-1, 1]$ based on their operational ranges.
To mitigate the possibility of simulation instability during the exploration process, the agent has been programmed to regulate the rate of valve opening and closing rather than controlling the opening percentage.
A hyperbolic tangent scales the designated action to the desired \ac{EGR} valve velocity with a range of $\pm \SI{50}{\percent\per\second}$.

\begin{table}[!t]
\caption{State, action, and reward quantities.}
\label{tab:agentIO}
\centering
{\renewcommand{\arraystretch}{1.2} 
\begin{tabularx}{\columnwidth}{@{}>{\hsize=0.3\hsize}X >{\hsize=0.7\hsize}X @{}}
\toprule
Type & Signal \\
\midrule
      & Engine speed $n_{t}$\\
      & Speed variation $n_{t} - n_{t - 1}$\\
      & Boost pressure $p_{act}$\\
      & Boost pressure variation $p_{act,t} - p_{act,t - 1}$\\
      & Boost target $p_{des}$\\
      & Boost target variation $p_{des,t} - p_{des,t - 1}$\\
State & Pedal position $Ped_{t}$\\
      & Pedal position variation $Ped_{t} - Ped_{t - 1}$\\
      & Boost control error $\Delta_{p}$\\
      & Coolant temperature $T_{cool}$\\
      & Gear number $i_{G}$\\
      & Vehicle speed $V_{veh}$\\
      & \ac{EGR} valve position $\phi$\\
\midrule
Action & Desired \ac{EGR} valve velocity $\omega_{\text{des}}$\\
\midrule
      & Engine-out NO\textsubscript{x} mass $m_{\text{NOx}}$\\
      & Engine-out Soot mass $m_{\text{soot}}$\\
Reward& Boost deviation $|p_{\text{des}} - p_{\text{act}}|$\\
      & Safety correction $\omega_{\text{des}}-\omega_{\text{safe}}$\\
      & Failure flag $b_{\text{fail}}$\\
\bottomrule
\end{tabularx}
} 
\end{table}

The reward function focuses on emissions formation (\emph{NO\textsubscript{x} and soot trade-off}), engine performance, combustion stability, and potential \ac{ECU} diagnostic trouble codes.
In both cases, a linearly negative reward is given to the emitted pollutant mass ($m\textsubscript{NOx}$ and $m\textsubscript{soot}$).
By penalizing the positive difference between the required and actual boost pressure ($p_{des}- p_{act}$ or $\Delta p$), an accurate torque provision is achieved.
Here, the boost error acts as a proxy since it is the only indicator for torque performance available in both \ac{MiL} and \ac{HiL} platforms.

Excessive \ac{EGR} rates during the vehicle's transition from idle to launch might lead to engine stalling, hence causing erroneous model performance.
Therefore, a safe, maximum \ac{EGR} position is defined during these phases and the agent's action is limited.
To guide the agent to avoid such behavior by itself, the difference between the agent's desired action and the safe one is penalized ($\omega_{\text{des}}-\omega_{\text{safe}}$ or $\Delta \omega$).
Lastly, given that the agent operates without internal algorithm constraints, the reward function incorporates an indicator to account for occurrences like model failures (e.g., simulation collapses due to divergence in internal physical calculations) and software failures in the \ac{ECU} ($F_{\text{fail}}$).
The resulting reward function is
\begin{equation} \label{eq:reward_func1}
\begin{split}
  r_{\text{XiL}} = & -f_{\text{NOx}} \cdot m_{\text{NOx}} - f_{\text{soot}} \cdot m_{\text{soot}} \\
  & - f_{\text{boost}} \cdot \Delta p \cdot 1\{\Delta p > 0\} - f_{\text{safe}} \cdot \Delta \omega - F_{\text{fail}}
\end{split}
\end{equation}

where $f\textsubscript{NOx}$, $f\textsubscript{soot}$, $f\textsubscript{boost}$ and $f\textsubscript{safe}$ are the weighting factors of NO\textsubscript{x} emissions, soot emissions, boost pressure deviation and action safety, respectively.
The reward function's weights were determined through tuning iterations, with the configuration from previous research serving as the starting point (table~\ref{tab:weights}).

\subsection{Software Architecture and Training Procedure}
\label{sec:XiL_Framework}
The connection of the agent to the plant models is achieved through the \ac{RL} I/O module, which consists of an interface to pre- and post-process relevant signals for the target application. 
The actions are communicated to the target function via automation and application software, ensuring efficient coordination throughout the system.
This configuration collects data from the plant models, processes signals, and computes the reward.

\begin{figure*}[t]
    \centering
    \includegraphics[scale=0.36]{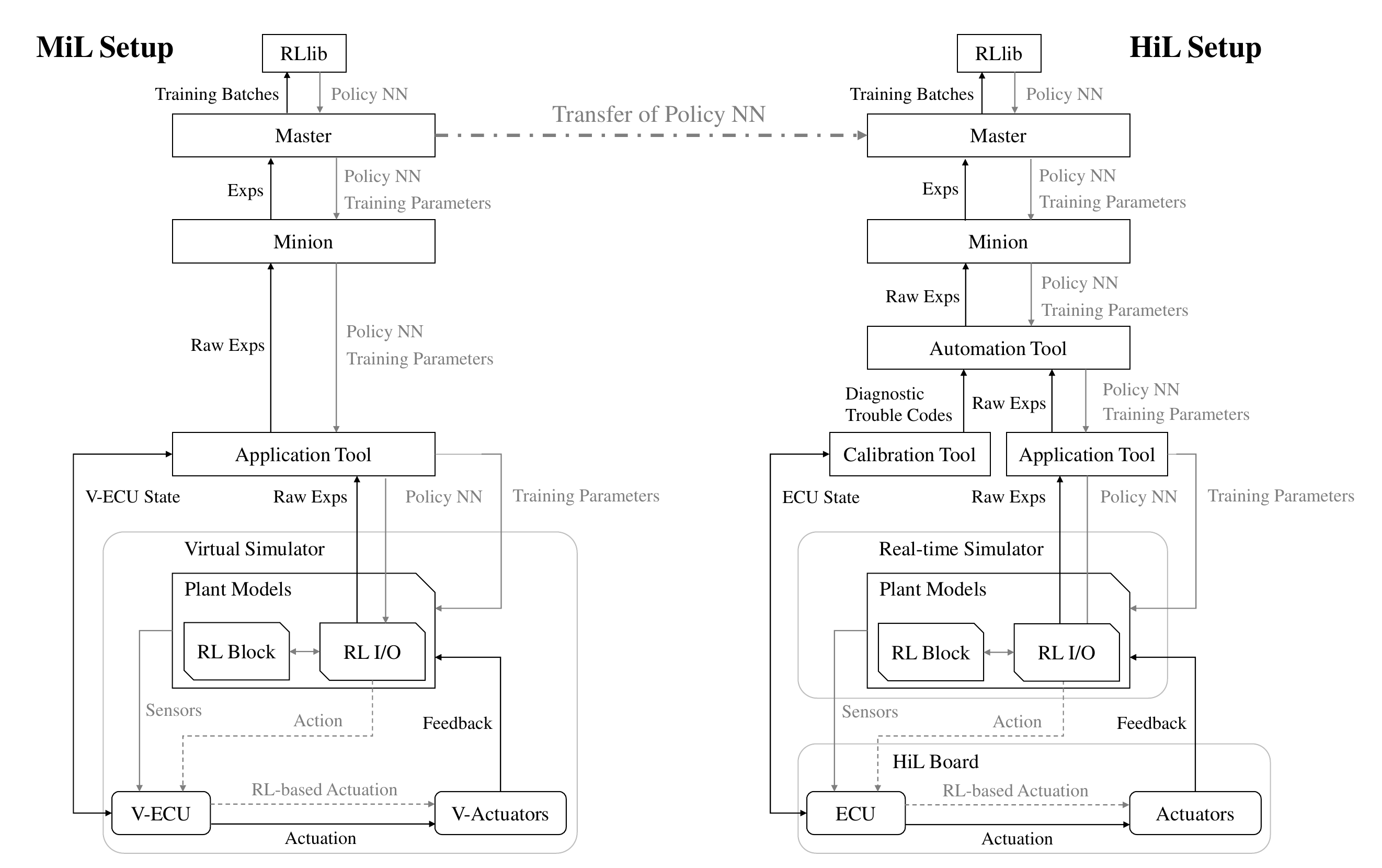}
    \caption{Framework architecture on \ac{MiL} (left) and \ac{HiL} (right), with training process based on \ac{TL}.}
    \label{fig:framework-setup-hil}
\end{figure*}

The policy network is represented by the network's edge weights and biases in a fully-connected feedforward \ac{NN} architecture, based on TensorFlow Lite for Microcontrollers\footnote{\url{https://www.tensorflow.org/lite/microcontrollers}}. It receives the processed states as input and generates an action distribution. From this distribution, an on-policy action is sampled to determine the agent's next action. The application software dynamically updates the network weights and biases before operation. 
The details and configuration of the \ac{MiL} training workflow have been described in previous work \cite{KOCH2023105477}.
On the \ac{HiL} platform, the agent adapts the default set points of the \ac{ECU}, which are then employed for controlling the real valve at each time step. A sample time of $t_{\text{s}}=\SI{0.2}{\second}$ has proven to be a good compromise between sample efficiency and control quality.

Training was conducted using a custom framework that enables on-policy algorithms to access remote embedded hardware for seamless coordination and communication between the necessary tools (Fig. \ref{fig:framework-setup-hil}).
Its first major component is the Master which interfaces the open-source library RLlib \cite{liang2018rllib}.
In addition to supervising the whole training process, the Master transfers the latest policy \ac{NN} and forwards gathered experiences to RLlib for training.
The second main element is the minion.
Master and minion use a custom protocol to exchange information via TCP/IP.
Because of their self-contained nature and the way they communicate, they can be located on different machines.
On the \ac{HiL} platform, the Minion is linked to the automation software, ECU-TEST\footnote{\url{https://www.tracetronic.com/products/ecu-test/}}.
The automation software, through Application Programming Interfaces, controls both the calibration software and the application software, namely ControlDesk\footnote{\url{https://www.dspace.com/en/pub/home/products/sw/experimentandvisualization/controldesk.cfm}}.
In this way, the minion has access to the real-time simulator and the plant models, where the agent runs in a special module (RL Block).
On the \ac{MiL} platform, the minion communicates directly with the application software, and consequently plant models, actuators (V-Actuators), and the model-based controller (V-ECU) in the virtual simulator.

The master initiates a training cycle by retrieving the current policy network and sending it to the minion along with a set of training parameters.
Through the control software, the Minion replaces the policy and sets the parameters in the plant models. After that, it engages in training episodes to collect a defined number of experiences. The Minion then retrieves the unprocessed experiences and transmits them to the Master, who stores them in a suitable format before passing them to RLlib for training. 
After a predefined number of training cycles, the Master runs a validation cycle. The difference between validation and training runs is that, during the former, actions are \emph{not} sampled stochastically from the action distribution, but instead the mean is taken. Thus, one obtains an unbiased and comparable evaluation of the agent's performance.

The policy transfer to the \ac{HiL} environment is handled by the Master.
The compatibility of the two environments in terms of state space representation and action space definition is assured, contributing to a smooth transition.
Through \emph{direct parameter transfer}, the network parameters
(including weights) obtained from the \ac{MiL} setup were replicated in the \ac{HiL} training.
It is reasonable to state that the agent executes the same task in domains that exhibit variations due to factors such as the impact of the real actuators, or software differences between the real controller and the simplified one used in \ac{MiL}.
However, Here, since \ac{HiL} and \ac{MiL} platforms use the same plant model, the agent only needed to fine-tune the pre-trained network, leading to a rapid adaptation.
Therefore a \emph{sample-efficient transfer} technique was employed for transferring it to the \ac{HiL} setup.

\begin{table}[!t]
\caption{Calibration of reward weights for \ac{MiL} training.}
\label{tab:weights}
\centering
\begin{tabularx}{\columnwidth}{@{}p{0.19\columnwidth}p{0.5\columnwidth}p{0.31\columnwidth}@{}}
\toprule
Weights & Description & Value \\
\midrule
$f_{\text{NOx}}$ & Factor for NO\textsubscript{x} emissions & \SI{0.14}{\per\gram} \\
$f_{\text{soot}}$ & Factor for soot emissions & \SI{1.42}{\per\gram} \\
$f_{\text{boost}}$ & Factor for boost deviation & \SI{0.00016}{\per\kilo\pascal} \\
$f_{\text{safe}}$ & Factor for action safety & \SI{0.015}{\second\per\percent} \\
$F_{\text{fail}}$ & Penalization for failures & \SI{20}{} \\ 
\bottomrule
\end{tabularx}
\end{table}

\section{Results}
\label{sec:Results}
This chapter presents the \ac{MiL} training, the policy transfer to \ac{HiL}, and the subsequent results when comparing the agents to the \ac{EGR} control function of the production \ac{ECU} in \ac{HiL}, and its model in \ac{MiL} which serve as references.

\subsection{Model-in-the-Loop Training}
The \ac{MiL} platform represents a safe and time-efficient source domain for \ac{TL}, where mature policies are generated within the order of hours depending on the selected training segments and computing hardware.
For this study, the training has been conducted on random segments of \SI{300}{\second} of a \ac{WLTC}, simulated with a real-time factor of \num{7.5} on a quad-core Intel Xeon W-\num{2123} processing unit.

In addition to generating transferable agents, the \ac{MiL} environment also serves as a platform for algorithm testing and hyperparameter tuning, which would be very time-demanding in the \ac{HiL} setup where training cannot be accelerated or scaled.
For the targeted use case, agents were trained with \ac{PPO} and \ac{DDPG} algorithms with equal definitions of states, action and reward (see Sec.~\ref{subsec:problem_formulation}).
The comparison of training progress and performance presented here does not aim to provide a comprehensive evaluation of individual \ac{RL} algorithms.
Rather, its purpose is to identify which of these promising algorithms is more suitable for the specific use case, considering control performance, time to convergence in training, and manual fine-tuning effort.
Tab.~\ref{tab:hyperparams} in Sec. \ref{sec:appendix1} lists the hyperparameter settings for both algorithms determined in the \ac{MiL} simulation.
The hyperparameter configuration utilized in the \ac{PPO} method is drawn from prior research.
Based on that, the hyperparameters of the \ac{DDPG} algorithm were selected and adjusted to achieve a stable training procedure that regularly outperforms the reference controller.

For both \ac{DDPG} and \ac{PPO}, the training was repeated three times with the same settings, but a random seed.
Tab.~\ref{tab:algorithm_comparison} shows the results of each algorithm's training and the performance of the reference controller.
The left side of the table contains characteristic parameters for evaluating the training progress based on the reward earned during each of the four training sessions.
On the right side of the table are use case-specific performance metrics for the agent with the best performance among the four training runs.
In comparison to \ac{PPO}, the \ac{DDPG} algorithm shows an increased reliance on hyperparameters and decreased reproducibility when using a fixed parameter set.
This is reflected by substantial deviations in the maximum reward of each training and a tripling of the average range of the validation reward over the course of each training cycle.
For the same training samples generated in each cycle, \ac{PPO} agents also showed quicker training progress.
The convergence time refers to the number of training cycles needed for the validation reward to be within \SI{5}{\percent} of the highest training reward.
In the case of \ac{DDPG} training, this time is significantly higher.

The driving cycle performance of the agent in terms of pollutant emissions and engine torque deviation is another indication of effective training.
For the emissions, the highest validation reward was achieved by a \ac{PPO} agent with comparable soot, but significantly lower NO\textsubscript{X} emissions.
Likewise, engine performance, as measured by the average absolute difference between target and actual engine torque, deviates only slightly. 
In comparison to the conventional reference controller, both \ac{RL}-based strategies exhibited improved performance across all three metrics.
Both the \ac{PPO}- and \ac{DDPG}-based agents managed to operate the engine without any monitoring interventions or model failures.

In summary, despite its off-policy nature, the \ac{DDPG} algorithm is found to require longer training time for the given use case.
While this aspect is acceptable in the context of \ac{MiL} simulations, it is a decisive disadvantage when training on real hardware.
Further, the \ac{PPO} algorithm demonstrates higher stability, which is crucial in \ac{HiL} setups where training repetitions are time-consuming, and most importantly, \ac{PPO} agents showed better overall performance compared to \ac{DDPG}.
Consequently, the \ac{PPO} algorithm has been chosen as the preferred approach for the \ac{HiL} transfer and will be the sole focus of the upcoming results.

\begin{table*}[t]
\caption{\ac{PPO} and \ac{DDPG} algorithms performance on the \ac{MiL} platform relative to the reference controller.}\label{tab:algorithm_comparison}
\begin{tabular*}{\textwidth}{@{\extracolsep{\fill}}ccccccccc}
\toprule
 & Std. deviation & Average & Avg. cycles & Max. & Rel. $m_{NO\textsubscript{X}}$ & Rel. $m_{Soot}$ & Rel. Avg $\Delta p$ & Rel. Avg $|\Delta V_{veh}|$\\
 & of max. reward & reward range & to convergence & reward & / \SI{}{\percent} & / \SI{}{\percent} & / \SI{}{\percent} & / \SI{}{\percent}\\
\midrule
\ac{DDPG} & \num{0.3776} & \num{5.049} & \num{113.8} & \num{-13.30} & \num{-2.39} & \num{-28.34} & \num{-80.71} & \num{-5.39}\\
\ac{PPO} & \num{0.09294} & \num{1.458} & \num{36.25} & \num{-12.71} & \num{-7.69} & \num{-26.45} & \num{-73.62} & \num{-3.95}\\
\bottomrule
\end{tabular*}
\end{table*}

\begin{figure}[h]
    \centering
    \includegraphics[scale=0.6]{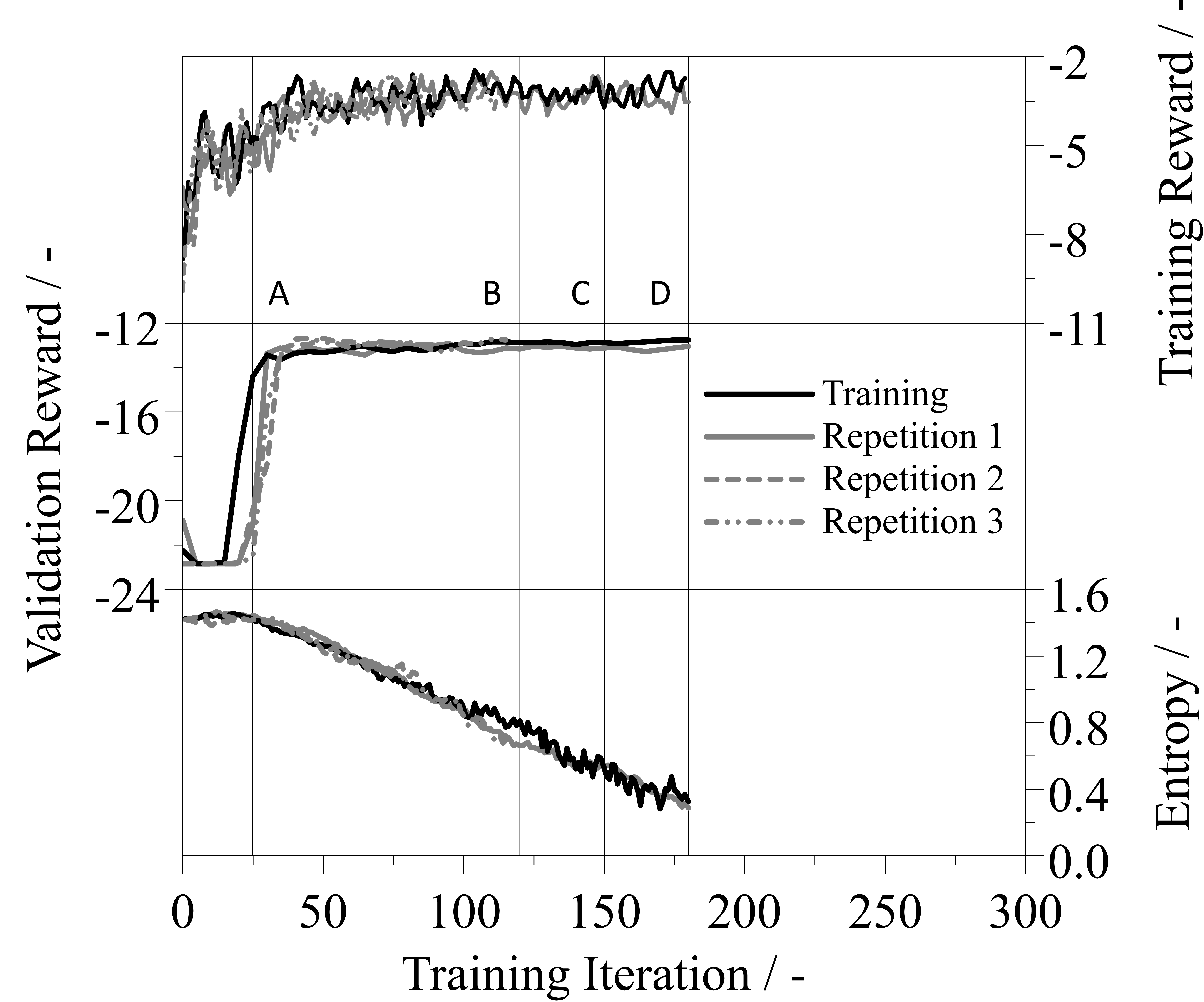}
    \caption{Progress, repetitions and transferred agents (A-D) in \ac{MiL} phase during training with \ac{PPO}.}
    \label{fig:MiL_training}
\end{figure}

The mean training reward, the validation reward, and the entropy of the four \ac{PPO}-based training runs analyzed in Tab.~\ref{tab:algorithm_comparison} are depicted in Fig.~\ref{fig:MiL_training} to demonstrate the repeatability of the training.
For all four training sessions, a similar, converging trend is observed in both training and validation rewards.
The fluctuations in the training reward can be attributed to the stochastic policy and the randomized segments of the driving cycle, some of which comprise primarily idling and others of high-speed sections that entail higher emissions.
The validation rewards that give a more unbiased evaluation of the agents' policies reveal a strong performance increase consistently after around \num{40} cycles for all of the agents until they converge at a similar maximum reward.
The decreasing entropy in the bottom plot confirms the growing maturity of the agents as the stochastic proportion of the policy and thereby the exploration drops continuously.

For the upcoming \ac{TL} investigations, four agents with different maturity levels are chosen.
The agents from the training marked in black in Fig.\ref{fig:MiL_training}, referred to as A (after \num{25} training iterations), B (\num{120}), C (\num{150}), and D (\num{180}) are considered potential candidates for \ac{TL} assessments.

\subsection{Hardware-in-the-Loop Training}
The investigations performed when transferring the \ac{PPO} training from the \ac{MiL} environment to the \ac{HiL} test bench are depicted in the following section.

\subsubsection{Fine-Tuning of Reward Parameters}
\begin{figure}
    \centering
    \includegraphics[scale=0.6]{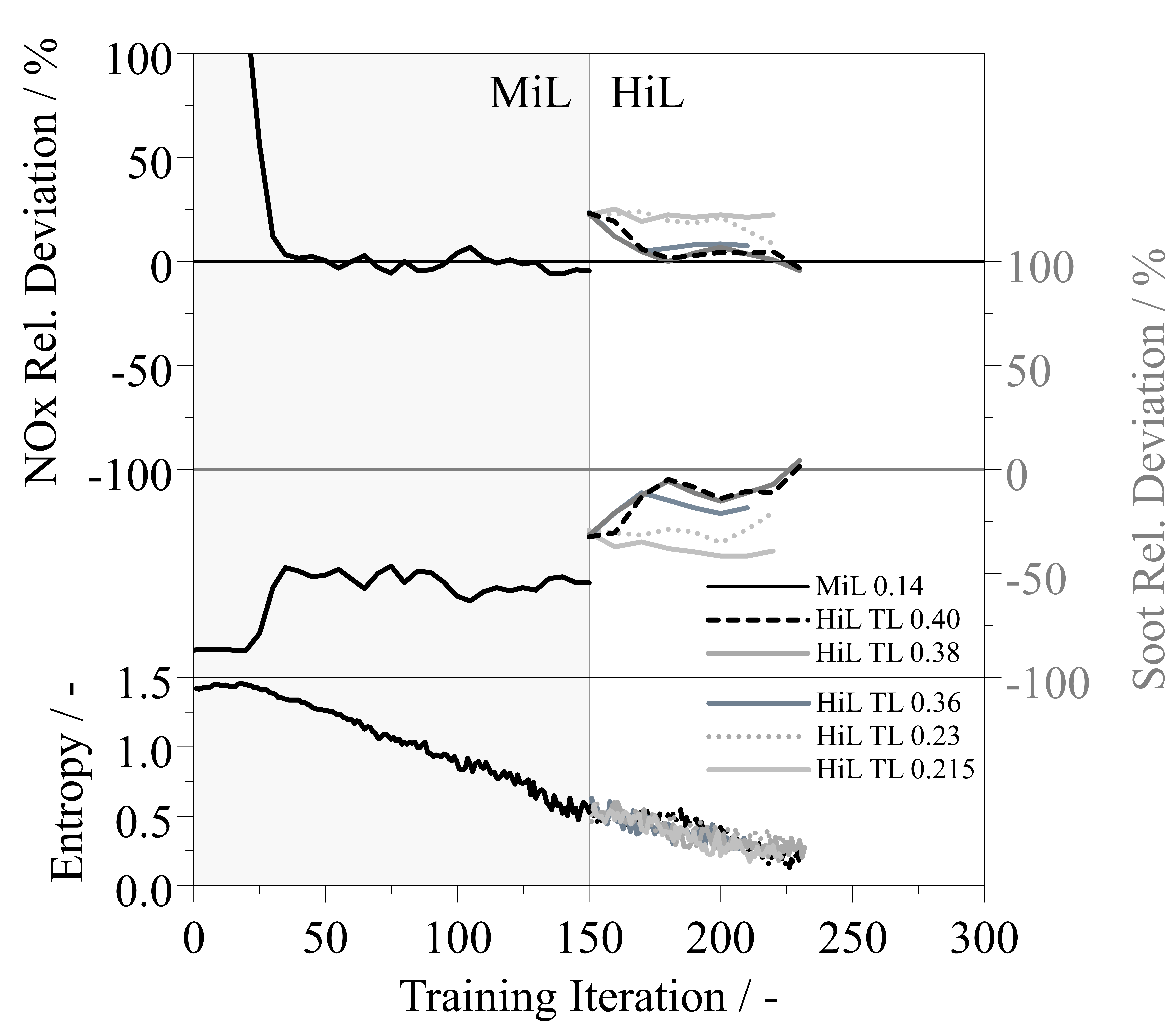}
    \caption{Impact of NO\textsubscript{X} reward factor variation based on \ac{WLTC} after transfer on the \ac{HiL} from \ac{MiL} after \num{150} cycles (agent C).}
    \label{fig:TL_1}
\end{figure}

Among all the options available, \emph{Agent C} (Policy \num{150}), which underwent \ac{PPO}-based \ac{MiL} training, exhibited notable progress in reducing emissions, so it has been chosen for the \ac{HiL} transfer.
However, when validating this policy to the \ac{HiL} without prior training, it did not balance emissions effectively (23\% reduction in NO\textsubscript{x} but a 30\% increase in soot emissions, compared to the reference).
Even though additional \ac{HiL} training increased the training reward marginally, it had no positive effect on the emission balance.
This indicates that the emission weights of the reward factors must be adjusted prior to training the agent in the a\ac{HiL} environment.
Starting the fine-tuning from scratch would necessitate delaying the evaluation of results until the strategy is completely developed.
Therefore, utilizing an agent who has already undergone extensive training can be advantageous.
Leveraging the mature policy of \emph{Agent C}, a sensitivity analysis is conducted by continuing the training on the \ac{HiL} from this checkpoint with different reward factor settings to determine the optimal calibration of the reward function that yields a balanced emissions trade-off.
Fig. \ref{fig:TL_1} shows the cumulative emissions and entropy trend for the \ac{MiL} and \ac{HiL} training progresses.
A total of five \ac{HiL} training runs were performed with the NO\textsubscript{x} reward as one degree of freedom while the other reward factors were kept as in the \ac{MiL} training.
By increasing the NO\textsubscript{x} penalty from \SI{0.215}{\second\per\gram} to \SI{0.40}{\second\per\gram}, the NO\textsubscript{x} emissions decreases as expected.
In this way, the emissions trade-off has been shifted below the reference \ac{ECU}, under the constraint of maintaining a reasonable engine performance. The results reveal a consistent decrease in entropy during \ac{TL} as a key outcome that underlines the robustness of the process.

In general, the \ac{HiL} training exhibits a greater cumulative deviation since it is susceptible to effects like latency and imperfect valve actuation from the real system.
Additionally, here the agent receives higher penalties for NO\textsubscript{x} emissions, which are also the most significant in terms of mass.
These results demonstrate that the selection of training environment and reward function is specific to the performance in the original environment.
The test with \SI{0.38}{\second\per\gram} for the NO\textsubscript{x} reward factor showed the best performances, and therefore it will be considered for further training with \emph{Agents A-D} on the \ac{HiL}.

\subsubsection{Impact of Transfer Learning on Performance}
After adapting the reward parameters, training is executed on the \ac{HiL} by transferring the four agents with varying maturity levels obtained from the \ac{MiL} training (agents A, B, C, D highlighted in Fig. \ref{fig:MiL_training}).
Furthermore, the outcomes of a pure \ac{HiL} training without previous policy transfer are also documented.

\begin{table*}[!b]
\caption{Comparison of \ac{TL} performance at different maturity levels and pure training on the \ac{HiL} platform.}
\label{tab:TLmaturity}
\small 
\begin{tabular*}{\textwidth}{@{\extracolsep{\fill}}cccccccc}
\toprule
Agent & \ac{MiL} training & \ac{HiL} training to max. reward & Max. & Rel. $m_{NO\textsubscript{X}}$ & Rel. $m_{Soot}$ & Rel. Avg $\Delta p$ & Rel. Avg $|\Delta V_{veh}|$\\
&/ iterations & / iterations (hours) & reward &/ \% &/ \% &/ \% &/ \% \\
\midrule
A & \num{25}   & \SI{185}{} (\SI{205}{})     & \SI{-27.98}{}  & \SI{-2.20}{}  & \SI{-1.50}{} & \SI{17.47}{} & \textbf{\SI{-10.89}{}}\\
B & \num{120}  & \SI{140}{} (\SI{156}{})             & \textbf{\SI{-27.66}{}}  & \textbf{\SI{-2.54}{}}  & \SI{-5.91}{} & \SI{4.37}{}  & \SI{-8.56}{}\\
C & \num{150}  & \textbf{\SI{30}{}} (\textbf{\SI{32}{}})      & \SI{-27.86}{}  & \SI{-0.10}{}  & \SI{-5.82}{} & \SI{2.90}{}  & \SI{-9.95}{}\\
D & \num{180}  & \SI{120}{} (\SI{134}{})             & \SI{-27.69}{}  & \SI{-0.96}{}  & \textbf{\SI{-8.97}{}} & \textbf{\SI{-0.04}{}} & \SI{-9.82}{}\\
Pure \ac{HiL}& \num{0} & \SI{170}{} (\SI{189}{})  & \SI{-28.31}{}  & \SI{-1.2}{}   & \SI{-0.1}{}  & \SI{15.88}{} & \SI{-7.65}{}\\
\bottomrule
\end{tabular*}
\end{table*}

The results shown in Tab.~\ref{tab:TLmaturity} present the maximum reward achieved and the corresponding training time in \ac{HiL} required to reach it.
Additionally, the relative emissions, boost pressure, and vehicle speed deviations are included, with the most beneficial values being visually highlighted.
From the table, it is evident that \emph{Agent C} stands out as the most time-efficient, as it achieved maximum reward after only \num{32} hours of \ac{HiL} training, showcasing its rapid convergence.
However, it is crucial to emphasize that the maximum reward does not necessarily equate to the best agent overall.
Examining the results, we note that some cases exhibit higher average boost pressure deviations $\Delta p$ compared to the reference; however, these deviations are minimal, merely in decimals.
These minor deviations do not impact vehicle speed performances.
The latter achieves an average difference of \SI{0.47}{\kilo\meter\per\hour} between the desired and actual speed when compared to the reference controller, with the corresponding relative values for the agents in the table being lower.
\emph{Agent A} performs well in terms of emissions and boost error, but it falls short in terms of total training duration.
For the specific use case, this demonstrates that the transfer of an agent with inadequate \ac{MiL} training is adverse to transferring a mature agent.
Finally, \emph{Agent C} stands out as a best-practice choice particularly for training efficacy in terms of time, but all agents have performance strengths.

\begin{figure}
    \centering
    \includegraphics[scale=0.6]{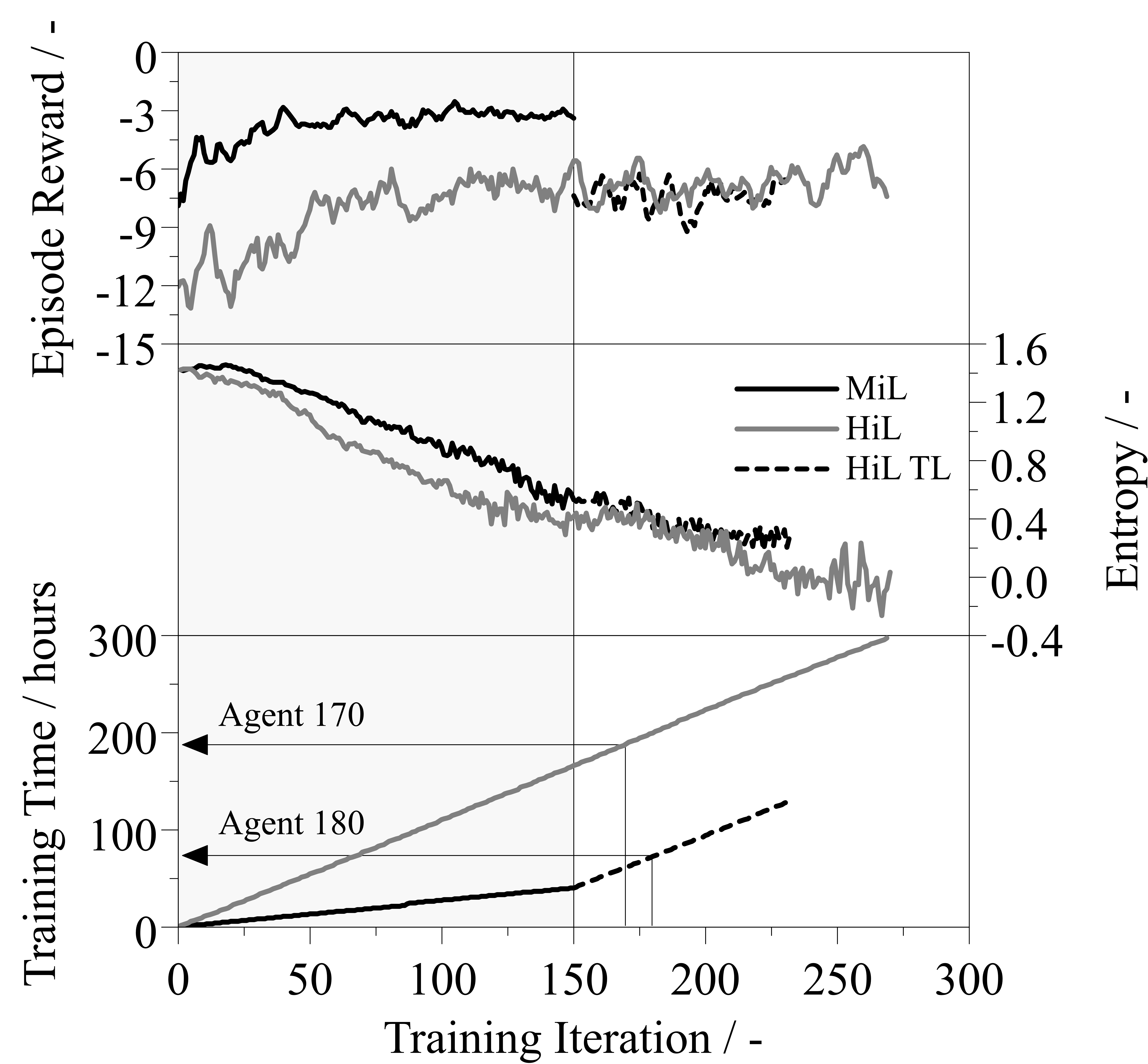}
    \caption{Full \ac{MiL}, \ac{TL}, and full \ac{HiL} training based on the \ac{WLTC}.}
    \label{fig:TL_3}
\end{figure}

Fig. \ref{fig:TL_3} depicts the episode reward, the entropy, and the equivalent training time for the \ac{MiL} training, the best-performing agent from Tab.~\ref{tab:TLmaturity} (\emph{Agent C}), and a complete \ac{HiL} training.
The training exhibits consistent mean episode rewards between transferred and pure \ac{HiL} agents.
The patterns observed in the entropy indicate that the \ac{HiL} training generally yields lower values.
When observing the equivalent training time, the curve representing the purely \ac{HiL} training shows that the best agent is attained after \num{189} hours (\emph{Agent \num{170}}).
In contrast, the best transferred agent is achieved after \num{72} hours (\emph{Agent \num{180}} with \num{40} hours in \ac{MiL} and \num{32} hours in \ac{HiL}), highlighting the substantial time savings in \ac{HiL} operation, which is \num{5.9} times shorter when applying \ac{TL}.

\subsubsection{Strategies Validation in \ac{HiL} Simulations}
\begin{figure}[ht]
    \centering
    \includegraphics[scale=0.38]{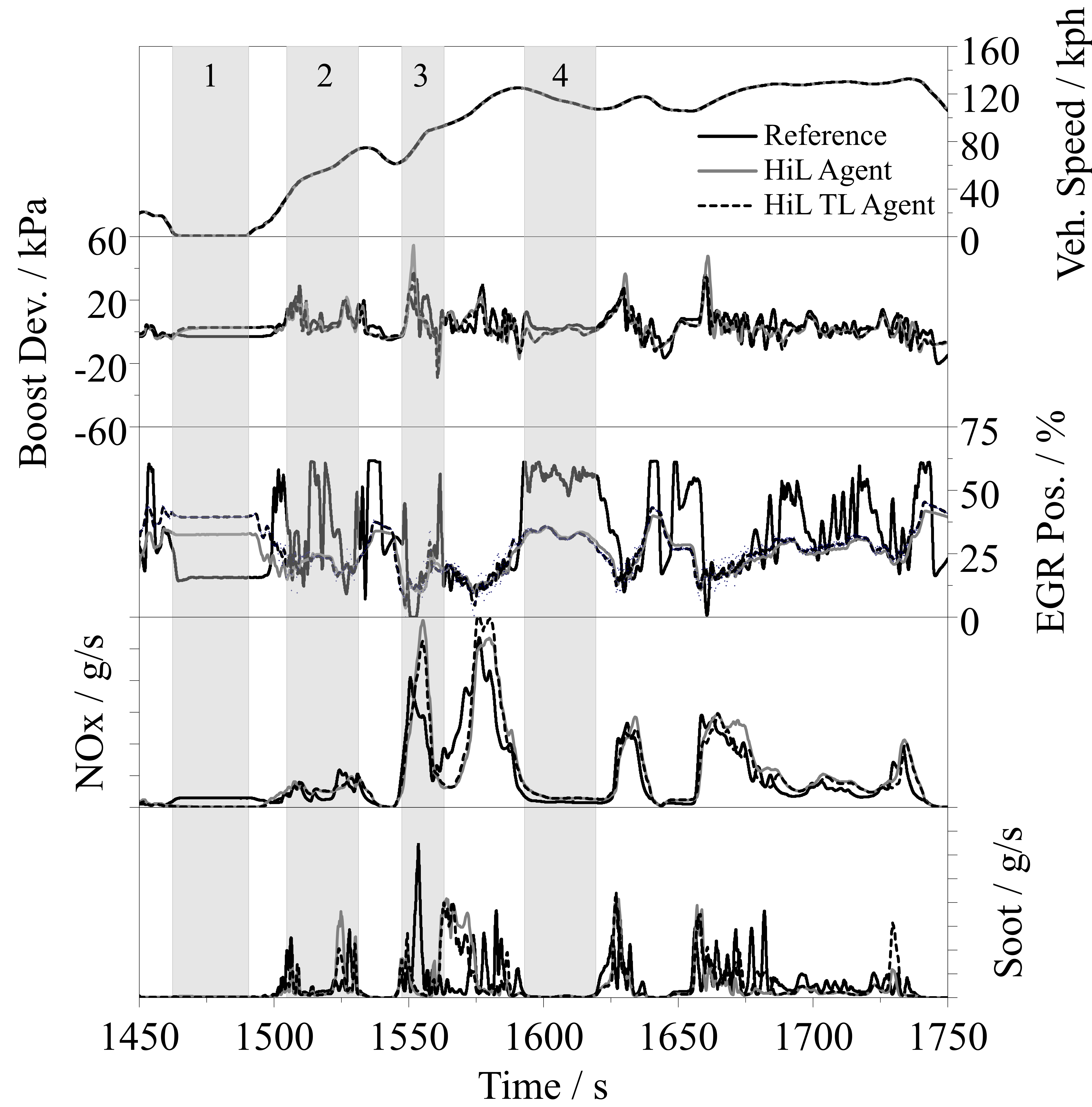}
    \caption{Comparison of control behaviors in four significant phases of the \ac{WLTC}.}
    \label{fig:HIL_1}
\end{figure}

A transferred agent and a pure \ac{HiL}-trained agent are compared to assess the performance when controlling real actuators in conjunction with the real \ac{ECU}. 
Fig. \ref{fig:HIL_1} shows vehicle speed, boost pressure deviation, \ac{EGR} valve position, and relevant emissions in function of the time during a \ac{WLTC}.
Four areas are highlighted: an idle phase (Area \num{1} at around \SI{1475}{\second}), two acceleration phases with dynamic actuation of the valves (Area \num{2} and Area \num{3} with up-shifting phases after \SI{1500}{\second}, and \SI{1550}{\second}), and a coasting phase (Area \num{4} at around \SI{1600}{\second}).
Observing the vehicle speed, it is evident that both agents control the vehicle without deviating from the desired speed, particularly at vehicle launch.
In general, the \emph{HiL} agent has a wider boost pressure deviation during up-shifting maneuvers, while both agents have higher deviation during idle phases.
However, this does not affect cycle performance and the average deviation over the cycle does not exceed \SI{0.6}{\percent} for both agents, compared to the reference controller.

The \ac{EGR} valve position from the \emph{HiL} and \emph{HiL TL} agents reaches a maximum of \SI{59}{\percent}, and \SI{51}{\percent}, respectively.
In general, the devised strategies are less dynamic without affecting vehicle performance.
The agents learned to request more \ac{EGR} during idle phases because it did not influence launching maneuvers from idle.
In Areas \num{2} and \num{4}, they are more conservative with \ac{EGR} than the reference controller and favor soot emissions reduction over NO\textsubscript{x} emissions.
Unlike the series \ac{ECU}, in Area \num{3} both agents request higher \ac{EGR} to reduce the soot peak emissions generated when up-shifting at high engine speeds.
Finally, with or without pre-training in \ac{MiL}, the agents were consistently able to outperform the reference controller in terms of cumulative fuel consumption and pollutant emissions.

\section{Conclusion}
\label{sec:conclusion}
In this paper, we have addressed the integration of \ac{RL} for the control development of embedded systems, specifically focusing on automotive applications.
Using a robust pipeline where \ac{RL} agents are trained and validated across multiple domains, we provide a technique that leverages varying levels of virtualization to enhance training efficiency by relocating agents from simulated environments to real-world situations.
The resulting agents, pre-trained in \ac{MiL} and trained on \ac{HiL}, outperform a series \ac{ECU} with minimal training duration.

At first, two model-free \ac{RL} algorithms, \ac{DDPG} and \ac{PPO}, were evaluated in the \ac{MiL} phase.
Despite its on-policy nature, which is associated with reduced sample efficiency, the \ac{PPO} algorithm displayed better performance and demanded comparatively less training time than the \ac{DDPG} algorithm.
From a pragmatic perspective, it additionally demonstrated improved reproducibility and reduced reliance on hyperparameters, rendering it a strong contender for few-shot transfer on a \ac{HiL} system with real hardware.

For the specific use case of \ac{EGR} control, we have investigated the implications of \ac{TL}, including the optimization of reward functions, the required maturity level prior to transfer, and a comparative analysis with cases without \ac{TL}.
The zero-shot transfer of \ac{MiL} agents revealed that the domain differences between the virtual and real \ac{ECU} require further optimization of the reward function.
The acceleration of this essential fine-tuning process was achieved by employing pre-trained agents rather than conducting training from the beginning.
When transferring agents with different maturity levels, the further \ac{HiL} training shows that fully-trained \ac{MiL} agents exhibit superior performance compared to ones that have not converged in terms of rewards and entropy. This superiority is observed not only in terms of training duration, but also in terms of control performance, as measured by maximum reward, relative emissions, and boost error reduction.
Additionally, an agent that was exclusively trained from scratch on the \ac{HiL} platform was compared against a transferred one. 
While the agent fully trained on \ac{HiL} managed to outperform the reference \ac{ECU}, achieving comparable emission reduction would have demanded extended training time on the real hardware.
The findings prove that the combination of \ac{MiL} pre-training with \ac{TL} improves the training process, enabling the attainment of a high-performing agent in much shorter time (\num{72} vs. \num{189} hours), while achieving up to \SI{5}{\percent} emissions reduction.
Ultimately, whether pre-trained in \ac{MiL} or not, all agents consistently surpassed the reference controller, with additional fuel savings up to \SI{3}{\percent}.


\section*{Acknowledgment}
The authors would like to acknowledge Norbert Meyer from dSPACE GmbH for his valuable support in the Vision project, as well as the relevant support team from TraceTronic GmbH for assistance with ECU-TEST.
Additionally, they acknowledge the valuable contributions of their student collaborators, including A. Hashmy and D. Antala.

\appendices

\section{Hyperparameters}
\label{sec:appendix1}
Tab.~\ref{tab:hyperparams} shows the hyperparameters settings used for the \ac{PPO} and \ac{DDPG} training.
\begin{table}[!t]
\caption{Hyperparameter settings for \ac{PPO} and \ac{DDPG}.}
\label{tab:hyperparams}
\centering
\begin{tabular}{lll}
\toprule
Hyperparameter & \ac{PPO} & \ac{DDPG}\\
\midrule
Policy \ac{NN} layers & \num{3} & \num{3} \\
Policy \ac{NN} layer size & \num{16} & \num{16} \\
Value f. \ac{NN} layers & \num{3} & \num{2} \\
Value f. \ac{NN} layer size & \num{16} & \num{80} \\
Activation function & hyp. tangent & hyp. tangent\\
Experiences per cycle & \num{9200} & \num{9200} \\
Train batch size & \num{9200} & \num{300}\\
Exploration method & sampling action distr. & Gaussian noise \\
SGD$^*$ size & \num{256} & -\\
Number of SGD steps & \num{32} & -\\
Learning rate $\eta$ & \num{0.00001} & - \\
Actor learning rate $\eta_{\text{ac}}$ & - & \num{0.00005}\\
Critic learning rate $\eta_{\text{cr}}$ & - & \num{0.0001}\\
Discount factor $\gamma$ & \num{0.9} & \num{0.95} \\
Target update factor $\tau$ & - & \num{0.03}\\
Replay training per cycle & - & \num{150}\\
\ac{PPO} clip parameter $\epsilon$ & \num{0.2} & -\\
GAE parameter $\lambda$ & \num{1} & -\\
KL penalty coefficient $\beta$ & \num{0.2} & -\\
Value loss coefficient $c_{\text{VF}}$ & \num{1} & -\\
\bottomrule
\end{tabular}
\caption*{\hspace{-9em}*SGD: Stochastic Gradient Descent.}
\end{table}

\input{IEEE_Picerno_RiL.bbl}

\end{document}